\documentclass[twocolumn]{article}
\usepackage[utf8]{inputenc}
\usepackage[ruled,vlined]{algorithm2e}
\usepackage{xcolor}
\usepackage{caption}
\usepackage[utf8]{inputenc}
\usepackage{amsmath}
\usepackage[colorinlistoftodos]{todonotes}
\usepackage{multirow}
\usepackage{graphicx}
\usepackage{amsmath}
\usepackage{breqn}
\usepackage{subcaption}
\usepackage{float}
\usepackage{enumerate}

\restylefloat{table}
\usepackage[margin=1in]{geometry}
\usepackage[utf8]{inputenc}
\usepackage{enumitem}
\usepackage{verbatim}
\usepackage[font=small,labelfont=bf]{caption}
\usepackage{tabularx}
\usepackage{makecell}
\usepackage{bbm}
\usepackage{cleveref}
\usepackage{biblatex}
\usepackage{todonotes}
\usepackage{sectsty}
\usepackage{authblk}
\usepackage[acronym,nogroupskip,nopostdot]{glossaries}
\newacronym{osm}{OSM}{Open Street Map}
\newacronym{dbscan}{DBSCAN}{Density-based Spatial Clustering of Applications with Noise}
\newacronym{gdp}{GDP}{Gross Domestic Product}
\newacronym{naics}{NAICS}{North American Industry Classification System}
\newacronym{api}{API}{Application Programming Interface}
\newacronym{poi}{POI}{Point of Interest}

\makeglossaries 

\title{Changes in Commuter Behavior from \mbox{COVID-19} Lockdowns in the Atlanta Metropolitan Area}
\author[1]{\mbox{Tejas Santanam}}
\author[1]{\mbox{Anthony Trasatti}}
\author[1]{\mbox{Hanyu Zhang}}
\author[1]{\mbox{Connor Riley}}
\author[1]{\mbox{Pascal Van Hentenryck}}
\author[2]{\mbox{Ramayya Krishnan}}
\affil[1]{H. Milton Stewart School of Industrial and Systems Engineering, \mbox{Georgia Institute of Technology}}
\affil[2]{Heinz College, \mbox{Carnegie Mellon University}}

\date{\today}
\begin{document}

\maketitle


\begin{abstract}
    \textbf{This paper analyzes the impact of \mbox{COVID-19} related lockdowns in the Atlanta, Georgia metropolitan area by examining commuter patterns in three periods: prior to, during, and after the pandemic lockdown. A cellular phone location dataset is utilized in a novel pipeline to infer the home and work locations of thousands of users from the \gls{dbscan} algorithm. The coordinates derived from the clustering are put through a reverse geocoding process from which word embeddings are extracted in order to categorize the industry of each work place based on the workplace name and \gls{poi} mapping. Frequencies of commute from home locations to work locations are analyzed in and across all three time periods. Public health and economic factors are discussed to explain potential reasons for the observed changes in commuter patterns.}
\end{abstract}

\section{Introduction}
This paper is motivated by a desire to gain an understanding of the impact of \mbox{COVID-19} on work-related travel behaviors in the Atlanta metropolitan area. The paper analyzes and answers the following questions: 
\begin{enumerate}
\item Where do people in the Atlanta metropolitan area live and work? 
\item How did people's commuter travel behaviors change over the course of the \mbox{COVID-19} pandemic and its associated lockdowns?
\item How have different industries changed their work styles as a result of the \mbox{COVID-19} pandemic and its associated lockdowns?
\end{enumerate}

From a high-level perspective, this paper provides a novel methodology for answering the above questions that consists of the following steps:
\begin{enumerate}
    \item Divide cell phone location data to users active before, during, and after \mbox{COVID-19} lockdowns in the Atlanta area.
    \item Apply density-based clustering to identify home locations, work locations, and days present at each.
    \item Use reverse geocoding combined with a bag-of-words model to map locations to workplace addresses and industries.
    \item Analyze user commuter and work behaviors in relation to work industry before, during, and after \mbox{COVID-19} lockdowns in the Atlanta area.
\end{enumerate}

The paper is organized as follows. Section~\ref{sec:literature_review} first presents a short survey of related literature. Next, Section~\ref{sec:data} describes the cell phone location dataset and its characteristics. Section~\ref{sec:clustering} follows with details on the clustering procedure, while Section~\ref{sec:reverse_geocoding} delves into the reverse geocoding and address mapping step. Section~\ref{sec:results} presents the final results of the analysis and Section~\ref{sec:discussion} ends with a discussion of the results.

\section{Literature Review} \label{sec:literature_review}

There have been a few recent papers that use cell phone location, GPS data, and smart card data to understand urban mobility patterns. The use of cell phone location data to generate home and work trip pairs, as well as assign some trip purposes are explored by \textcite{ccolak2015analyzing}. \textcite{ahas2009modelling} compare Estonian mobile data with population register data to analyze the effectiveness of a model that determined home and work locations for users. More recently, \textcite{sari2019high} use smart card data for home and work location detection, while \textcite{yang2018understanding} use mobile data to understand the spatial structure of urban commuting. \textcite{xu2015understanding} also use mobile data to examine how people move around under the context of key activity locations. Patterns in \mbox{home-work} commuting are also examined by \textcite{kung2014exploring}. \textcite{zhang2020learning} focus on learning taxi driver behavior from GPS data.

The use of \mbox{density-based} clustering methods on mobile phone data has also been explored in prior literature. \textcite{jiang2022improved} improve on the existing \gls{dbscan} algorithm methods to identify trip ends from Chinese mobile phone location data. \textcite{zhang2021research} use DBSCAN clustering to cluster public transit passengers into regional travel areas. \textcite{sakamanee2020methods} implement DBSCAN for noise reduction as they infer route choice for commuter trips. DBSCAN is used as the final step in deriving the characteristics of commuting patterns on key commuting routes by \textcite{yao2022analysis}.

A few recent papers have looked at the impact of the \mbox{COVID-19} pandemic on human mobility patterns. \textcite{engle2020staying} estimate how individual mobility outside the home is affected by local disease prevalence and restriction orders to \mbox{stay-at-home}. \textcite{bick2021work} document the evolution of \mbox{work-from-home} behaviors over the course of the pandemic and the expectation for that kind of working behavior in the future. \textcite{chen2021nursing} uses cell phone location data to understand the impact of the pandemic on nursing home entries. \textcite{huang2022staying} look at the correlation between demographic and socioeconomic variables and \mbox{home-dwelling} time records as a result of \mbox{stay-at-home} orders derived from \mbox{large-scale} mobile phone location tracking data. \textcite{zhang2020interactive} build an interactive tool and process anonymized mobile device location data to identify trips and produce a set of variables including a social distancing index, the percentage of people staying at home, the number of visits to work and \mbox{non-work} locations, \mbox{out-of-town} trips, and trip distance. \textcite{jiao2021measuring} look at how the number of \mbox{COVID-19} cases can be used to predict the following week's travel behaviors in Houston, Texas. \textcite{pourfalatoun2022effects} analyze the impacts of \mbox{COVID-19} on micromobility use and perception.

The choice of the Atlanta metropolitan area as a case study in this paper is also based in prior literature analyzing location data, commuter behavior, and commuter location data. \textcite{schonfelder2006analysis} use GPS data to analyze destination choice behavior for vehicle trips in Atlanta. \textcite{santanam2021public} and \textcite{trasatti2022data} discuss the consistency of commuter behavior using public rail in Atlanta. The impact on this consistency in Atlanta as a result of the pandemic is analyzed by \textcite{auad2021resiliency} who also propose a resilient multimodal transit system to serve the Atlanta area in a pandemic case in the context of \mbox{COVID-19}. \textcite{guan2022heuristic} consider commuter home locations in the integration of latent demand into the design of an on-demand multimodal transit system for Atlanta. The use of the OpenStreetMap \gls{api} for Atlanta area points of interest has also been implemented by \textcite{li2022ease} who build a tool to explore the transit accessibility in the Atlanta metropolitan area to those points of interest.

Based on the literature above, there exists a gap in using cell phone (or any similar type of) location data to derive a breakdown of remote and in-person commute behaviors by work industry. Ultimately, the main novel contributions of this paper relate to the analysis of lockdown effects on commuter behaviors. To our knowledge, there are no comparable methods in the existing literature for deriving granular analysis on commuting patterns by industry from mobile location data. The pipeline that connects density-based clustering to reverse geocoding to work type is also a unique and novel contribution, and is scalable in its potential applications outside of the Atlanta area and in temporal periods beyond the focus on \mbox{COVID-19} lockdowns.

\section{Data} \label{sec:data}
\begin{figure}[!t]
\centering
\includegraphics [width=1\linewidth] {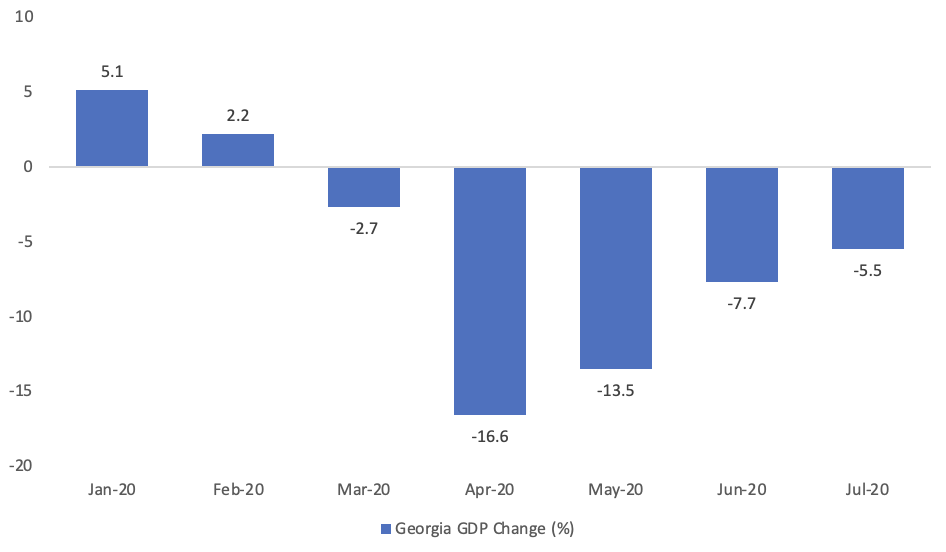}
\caption{Monthly change in Georgia GDP (early 2020)}
\label{fig:gdp}
\end{figure}

The paper uses a dataset from \mbox{X-Mode} which contains the cell phone latitude and longitude location data for a set of anonymized users who use mobile applications with the \mbox{X-Mode} platform integrated. X-Mode is a data collection platform for location data that app developers can choose to integrate into their app \cite{x_2022}. The original dataset contains over 12 terabytes of location data for users all over the United States, so the first step is to preprocess the data to focus on the target time and locations. The data is filtered and contained to a bounding box around the metropolitan area surrounding the city of Atlanta. The city of Atlanta is located in Fulton County, Georgia which is the county in Georgia that saw the largest number of \mbox{COVID-19} cases and related deaths in 2020, with 56,800 reported \mbox{COVID-19} cases over the year. The resulting data points are further filtered to include user locations during the months of January 2020 (baseline pre-pandemic), April 2020 (pandemic lockdown), and July 2020 (pandemic post-lockdown). The \gls{gdp} of the state of Georgia grew by 5.1\% in January 2020 compared to January 2019, decreased by 16.6\% in April 2020 compared to April 2019 (the largest decrease in the last decade), and fell by 5.5\% in July 2020 compared to July 2019. The monthly GDP trend is shown in Figure~\ref{fig:gdp}. These months are analyzed and focused on in this paper. 

\begin{figure}[!t]
\centering
\includegraphics [width=1\linewidth] {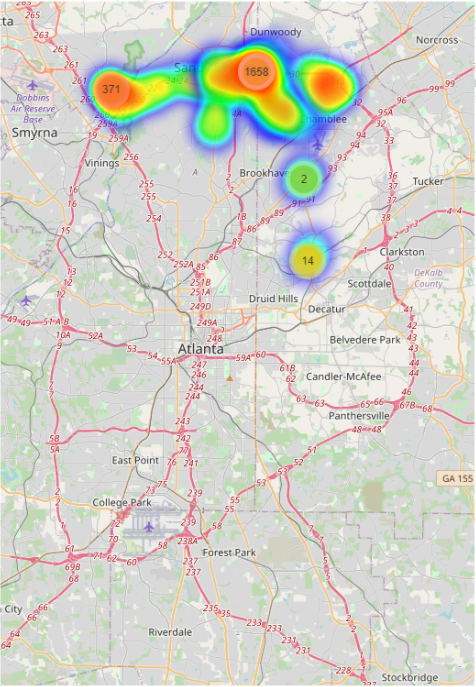}
\caption{Location data for one user over a month}
\label{fig:oneuserpts}
\end{figure}

In January 2020, which was used as a baseline, there were around 430,000 unique users who had at least one location point tracked and logged to the dataset. In order to achieve a generalization of the users' travel patterns, the number of active days for each user in January is calculated. A threshold of greater than 24 days with at least one data point present in the dataset is set to maintain a strong clustering result. With this threshold, a little over 21,000 users are identified as having enough information in the month of January. The location data for a single month for one anonymized user is displayed on the map as point clusters and a heat map in Figure~\ref{fig:oneuserpts}.

\section{Clustering Methods} \label{sec:clustering}

To begin the analysis of commuter behavior from the location data, work and home locations are needed to be determined. To do so, a clustering approach is used. Work location and home location for each user are identified using DBSCAN clustering. DBSCAN is a \mbox{density-based} clustering algorithm that takes a set of points in some space and groups together points that are closely packed together \cite{ester1996densitybased}. It is used here due to its quality performance in finding core samples of high density at home and work locations and expanding clusters from them. The following process of location identification below is used:
\begin{enumerate}
    \item Starting with the filtered data containing 21,000 users from above, further filter the data to the group of users with more than 300 data points of activity during the pre-pandemic period in January. If a user has fewer than 300 activities, the home and work locations derived might not be accurate. Those users with fewer than 300 data points of activity are not assigned any home and work locations. 4,500 of the 21,000 user subset meet this criteria.
    \item Taking the filtered group of users, divide the user’s GPS data into workdays and weekend. For each \mbox{sub-dataset}, cluster the GPS data using DBSCAN. The parameters for DBSCAN are set to be eps = 0.001, and minsample = 5\% of the sample size and were manually tuned. 
    \item Find the cluster centers for the above users for workdays and weekends.
    \item For each cluster center, find which days they appear in within a month.

    \item The cluster center that appears most frequently in terms of days in January (that also has a frequency greater than 17 days) is classified as the ``home location''. If there is no ``home location'' found, then go to the next user.
    \item Remove the ``home location''  from the list of cluster centers and find the most frequent location that appear in workdays in the list of cluster centers. If that location appears more than 10 days per month, then identify that location as ``work location''. 
\end{enumerate}

\begin{figure}[!t]
\centering
\includegraphics [width=1\linewidth] {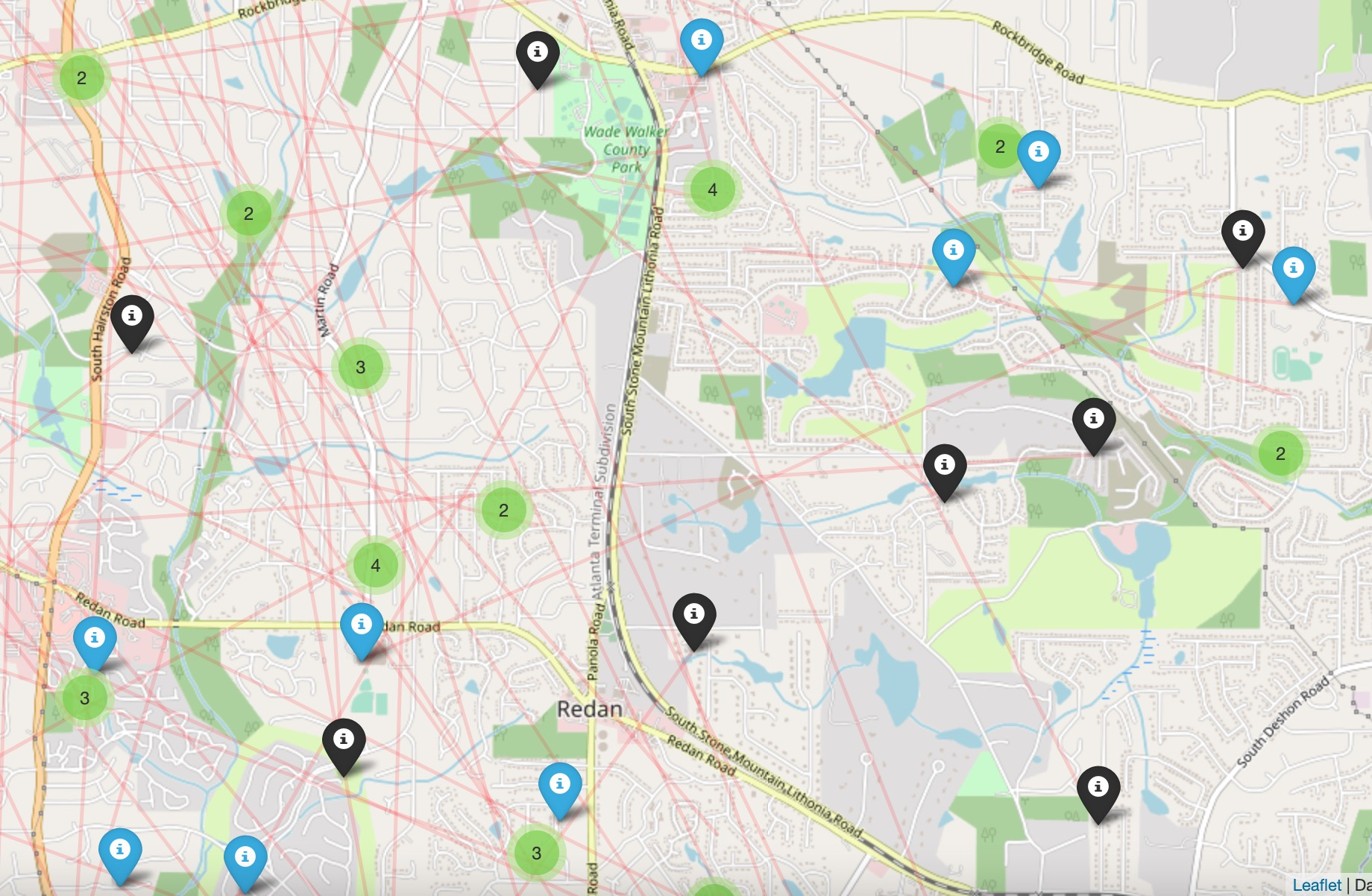}
\caption{Home and Work locations in the 1.5K user subset}
\label{fig:homeandworkplot}
\end{figure}

\begin{figure}[!t]
\centering
\includegraphics [width=1\linewidth] {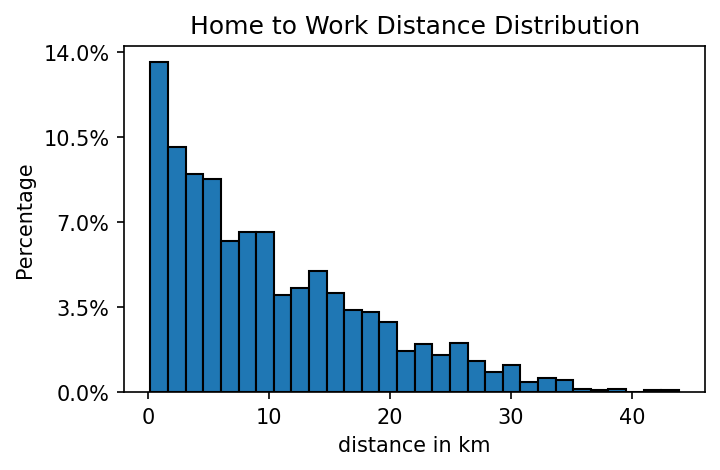}
\caption{Home to Work distances in the 1.5K user subset}
\label{fig:hometoworkdist}
\end{figure}

Using this method, the anonymous user in Figure~\ref{fig:oneuserpts} will have their home location identified within the top cluster of 1,658 points and their work location identified within the leftmost cluster of 371 points. After repeating the above steps on the 21,000 user subset, 1,428 users are also above the 24 day activity threshold in April and July as well. This roughly 1.5K user subset is the main focus of the analysis going forward in this paper. These clusters were also manually inspected to obtain a high-level of confidence in the accuracy of the home and work labels based on the general locations of where one might expect to see residential and commercial areas. The home locations (in black) and work locations (in blue) of some of those 1,428 users are shown as markers and marker clusters in Figure~\ref{fig:homeandworkplot}. Home and work locations for any particular user are connected in the figure by a red line. The distribution of the distances between the home and work locations for these users are shown in kilometers in Figure \ref{fig:hometoworkdist}. 
\section{Reverse Geocoding} \label{sec:reverse_geocoding}

To figure out the work types of these users and analyze the change in their work patterns during the pandemic, a reverse geocoding procedure is applied on the latitude and longitude of their work locations. The Nominatim OpenStreetMap API is used to retrieve the actual name of the work locations; for example, ``Drury Inn'' or ``Wells Fargo'' \cite{OpenStreetMap} \cite{clemens2015geocoding}. For the set of work locations where the Nominatim OpenStreetMap API is unable to provide a workplace name, a database of Atlanta points of interest is used to map the work locations to workplace names. For the remainder of the few unmapped work locations, a manual mapping is done through Google Maps. 

After the workplace names for all of the users are gathered, a bag-of-words model is run on the work place names. The purpose of the bag-of-words model is to count the frequency of the words that appear in the names of workplaces in the roughly 1.5K workplace set. Some high-frequency words are identified such as ``school'', ``hotel'', and ``hospital''. These high-frequency words serve as guidance for designing the word embeddings to extract categories from location names. The generated categories are presented in Table~\ref{tab:numcattype}. The word embeddings extract categories from the location names. For example, any workplace with the word ``University'', ``School'', or ``Academy`` in the title is mapped by the word embeddings program to the Education category (except in the case of workplace names like ``Emory University Hospital'' which is mapped to the Medical category). Similarly, workplace names containing ``Taco'', ``Pizza'', or ``Diner'' are mapped to the Restaurant category. 
\begin{figure}[!t]
\centering
\includegraphics [width=1\linewidth] {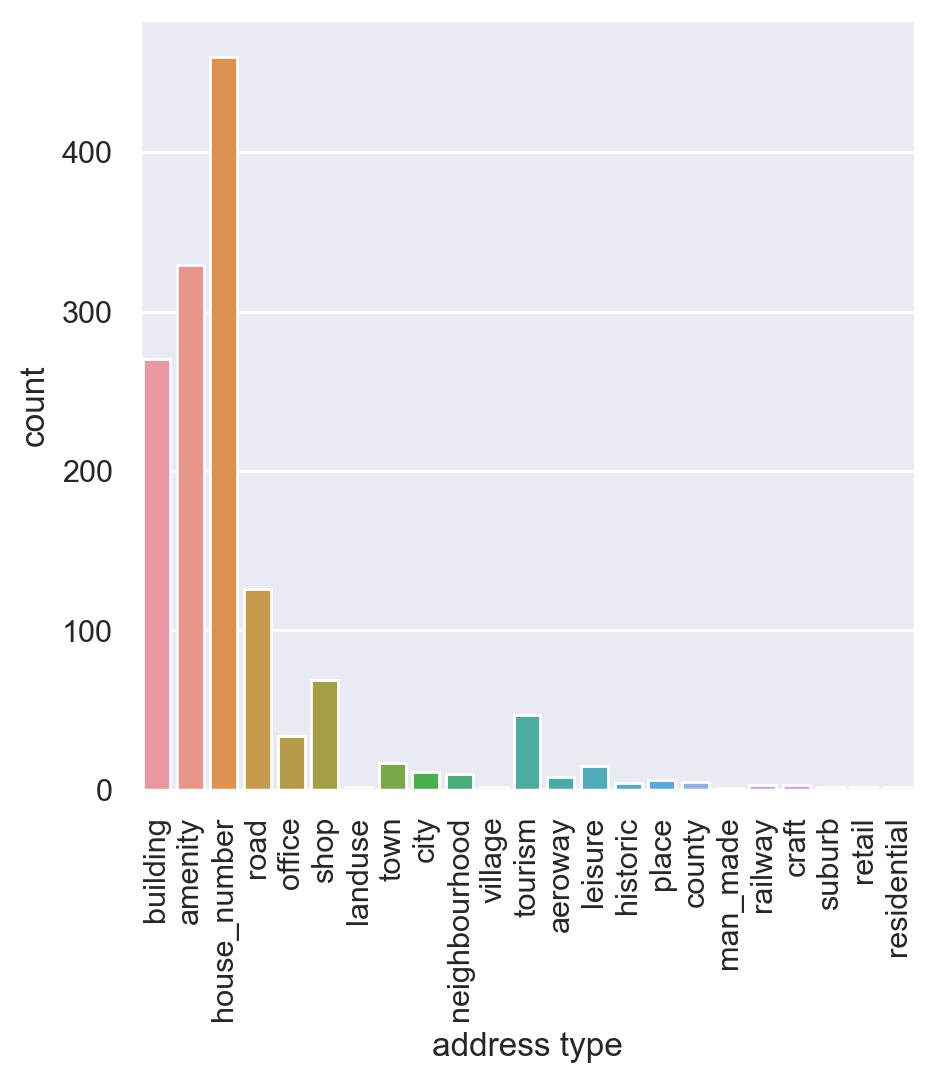}
\caption{Category of Nominatim OpenStreetMap API address matches}
\label{fig:apiaddress}
\end{figure}

\section{Results} \label{sec:results}
\begin{figure}[!t]
\centering
\includegraphics [width=1\linewidth] {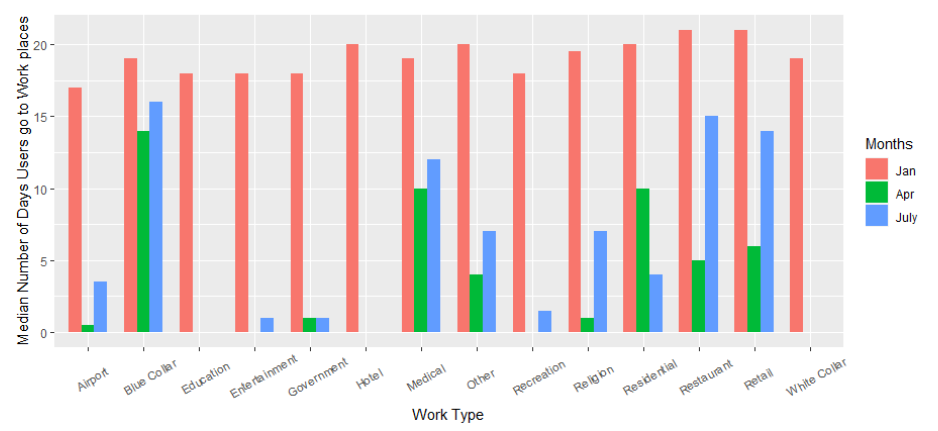}
\caption{Bar graph of median workdays by workplace category}
\label{fig:medworkdaybar}
\end{figure}

\begin{figure}[!t]
\centering
\includegraphics [width=1\linewidth] {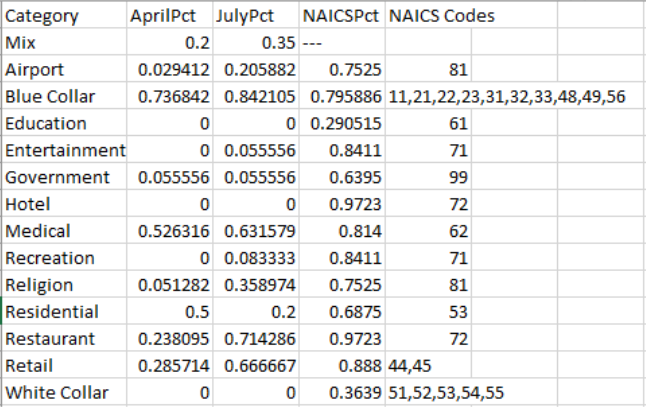}
\caption{Comparison of actual workday percentage to NAICS in-person work percentages}
\label{fig:naics}
\end{figure}

Table~\ref{tab:numcattype} displays the number of people associated with each workplace category out of the 1,428 users analyzed. The notes column details the types of workplaces that were mapped to each category at the reverse geocoding step. The largest categories contained people in Mix, Retail, White Collar, and Education jobs. A comparison between the 1,428 user subset and the roughly 21,000 users who were only active in January in the dataset (not including the users in the 1,428 user subset) is created, and it is found that the distribution of job type by percentage is roughly similar. This can verify that the subsets generated in the analysis are representative of the overall Atlanta metropolitan area community. The median workdays over each month and the associated percentiles are shown in Table~\ref{tab:medworkday}. In that table, the 25\% percentiles refers to fewer than 25\% of the users in that category working fewer than the specified number of days. The median number of days is graphed in Figure~\ref{fig:medworkdaybar}. While all occupation types see a decrease in the number of days users went in to work, the Blue Collar and Medical fields see the greatest resilience of \mbox{in-person work}. Some fields like the White Collar and Education sectors see their median day count drop to zero as a result of the possibility of virtual transition these sectors have. Some sectors like the hotel industry or entertainment venues also see a median day count drop to zero as a result of forced closings and layoffs from lockdown policies in Atlanta. Figure~\ref{fig:naics} shows a comparison between the percentage of days worked in person during the pandemic months compared to the expected percentage of days worked in person from the \gls{naics} dataset, which is a government dataset for businesses in Canada, Mexico, and the United States on how necessary \mbox{in-person} work is within different occupations and industries.

\begin{table}[!t]
\centering
\resizebox{\linewidth}{!}{%
\begin{tabular}{l|l|l}
Categories    & Count & Note                                          \\\hline
Airport       & 14             & Airport Terminals, Parking Lots                                     \\
Blue Collar   & 59             & \makecell[tl]{Manufacturing, Vehicle Repair, \\Cemetery Work}                        \\
Education     & 156            & K-12 schools and Higher education, Preschools                       \\
Entertainment & 9              & \makecell[tl]{Stadium, theatre, theater, movie, bowling,\\ opera, museum, orchestra} \\
Government    & 30             & City Halls, Mints, Consulates                                       \\
Hotel         & 51             & Hotels, Motels                                                      \\
Medical       & 130            & \makecell[tl]{Hospitals, Urgent Care, \\ Dentists/Chiropractic Offices}               \\
Mix           & 385            & Small Businesses, Home Businesses, etc.                             \\
Recreation    & 16             & Trails, Parks, Golf Courses, Tennis Courts                          \\
Religion      & 28             & Churches, Temples, Mosques                                          \\
Residential   & 70             & Apartments, Townhomes, Condos                                       \\
Restaurant    & 90             & Food and Drink eat-in Establishments                                \\
Retail        & 203            & \mbox{Brick-and-mortar} Retail                                             \\
White Collar  & 187            & Large Office Buildings/Companies                                   
\end{tabular}%
}
\caption{Amount of users by workplace category in the 1.5K user subset}
\label{tab:numcattype}
\end{table}


\begin{table}[!t]
\centering
\resizebox{\linewidth}{!}{%
\begin{tabular}{l|rrr|rrr|rrr}
\multicolumn{1}{l}{} & \multicolumn{3}{c}{Jan}           & \multicolumn{3}{c}{Apr}                                & \multicolumn{3}{c}{July}                                  \\
Categories    & 50\% & 25\% & 75\% & 50\% & 25\% & 75\% & 50\% &  25\% &  75\% \\ \hline
Other         & 20          & 17       & 22       & 4           & 0        & 17       & 7            & 0         & 19        \\
Airport       & 17          & 16       & 22     & 1         & 0        & 4        & 4          & 0         & 13        \\
Blue Collar   & 19          & 16       & 21       & 14          & 1        & 20       & 16           & 3       & 20        \\
Education     & 18          & 16       & 19       & 0           & 0        & 1        & 0            & 0         & 3         \\
Entertainment & 18          & 16       & 19       & 0           & 0        & 0        & 1            & 0         & 4         \\
Government    & 18          & 15       & 19       & 1           & 0        & 3     & 1            & 0         & 5      \\
Hotel         & 20          & 16       & 22       & 0           & 0        & 7        & 0            & 0         & 15      \\
Medical       & 19          & 16       & 21       & 10          & 1        & 19       & 12           & 2         & 18        \\
Recreation    & 18          & 14       & 22       & 0           & 0        & 3      & 2          & 0         & 18     \\
Religion      & 20        & 15       & 23    & 1           & 0        & 6      & 7            & 0         & 13        \\
Residential   & 20          & 17       & 23       & 10          & 0     & 22     & 4            & 0         & 22     \\
Restaurant    & 21          & 18       & 24       & 5           & 0        & 18    & 15           & 1         & 20     \\
Retail        & 21          & 18       & 24       & 6           & 0        & 20       & 14           & 1         & 20        \\
White Collar  & 19          & 16       & 21       & 0           & 0        & 2        & 0            & 0         & 5        
\end{tabular}%
}
\caption{Median and quartiles of workdays by workplace category}
\label{tab:medworkday}
\end{table}

\section{Discussion} \label{sec:discussion}
The most impactful conclusion stems from the three different types of median workday changes over time seen in the results. One type of trajectory is a sudden large drop in \mbox{in-person} work as the \mbox{COVID-19} pandemic struck that continues post-lockdown as well. This is seen in categories like White Collar, Education, Entertainment, Hotel, and Government. The second type of trajectory that is noticeable is a large drop as \mbox{COVID-19} struck that then bounces back (but not all the way to normal levels) \mbox{post-lockdown}. This trajectory is seen in categories like Restaurants and Retail locations. The third type of trajectory is a small drop to a level that stays consistent through lockdown and \mbox{post-lockdown}, as seen in Blue Collar and Medical work. For the most part, the first category contains jobs that either can easily be transitioned virtually or jobs that are impossible to transition virtually and thus had to pause. The second category contains those kinds of jobs that had to temporarily stop during lockdown but were able to come back in \mbox{non-traditional} ways such as outdoor seating-only for restaurants or online order and pick-up operations in retail locations. The third category contains those jobs that had to continue in-person regardless of the pandemic conditions, such as medical work or running a grocery store. Further conclusions can be drawn from the NAICS percentages. While some \mbox{in-person} work percentages align with the government expectation for \mbox{work-from-home} \mbox{pre-COVID-19}, as seen in the case of Blue Collar work, many occupations see a large drop from the expectation during lockdown, such as a 73\% drop in Restaurant work and Airport work or a 97\% drop in Hotel work. This large drop from the expectation can be attributed to the effect of legislation and policy on in-person work. Since legislation forced the closing of many \mbox{service-based} companies such as bowling alleys, theaters and restaurants that did not have outdoor seating, these jobs that would normally be considered by the government to be \mbox{in-person} were no longer possible. 
Potential areas of future work include expanding this analysis to other cities and comparing the results with Atlanta. Another interesting line of analysis would be to analyze these workplaces from the customer perspective and see where people went during \mbox{COVID-19}. Credit card data could be used to compare where consumers spent their money in-person versus online. The addition of smart card data and other forms of public transit data sources can also be used to further the granularity of the analysis over time in this paper. After some time, analysis of the \mbox{long-term} effects and implications of the pandemic on the greater commuter culture would be useful to explore.

\printbibliography[
heading=bibintoc,
title={References}
] 

\clearpage

\end{document}